\documentclass[10pt,twocolumn,letterpaper]{article}

\pdfoutput=1

\usepackage{cvpr}
\usepackage{float}
\usepackage{times}
\usepackage{epsfig}
\usepackage{graphicx}
\usepackage[caption = false]{subfig}
\usepackage{amsmath}
\usepackage{amssymb}
\graphicspath{{images/}}
\usepackage{caption} 
\usepackage[breaklinks=true,bookmarks=false]{hyperref}

\cvprfinalcopy % *** Uncomment this line for the final submission

\def\cvprPaperID{****} % *** Enter the CVPR Paper ID here

\begin{document}

%%%%%%%%% TITLE
\title{Panoptic Segmentation with a Joint Semantic \\ and Instance Segmentation Network}

\author{Daan de Geus\\
Eindhoven University of Technology\\
{\tt\small d.c.d.geus@tue.nl}
\and
Panagiotis Meletis\\
Eindhoven University of Technology\\
{\tt\small p.meletis@tue.nl}
\and
Gijs Dubbelman\\
Eindhoven University of Technology\\
{\tt\small g.dubbelman@tue.nl}
}

\maketitle
%\thispagestyle{empty}

%%%%%%%%% ABSTRACT
\begin{abstract}
We present a single network method for panoptic segmentation. This method combines the predictions from a jointly trained semantic and instance segmentation network using heuristics. Joint training is the first step towards an end-to-end panoptic segmentation network and is faster and more memory efficient than training and predicting with two networks, as done in previous work. The architecture consists of a ResNet-50 feature extractor shared by the semantic segmentation and instance segmentation branch. For instance segmentation, a Mask R-CNN type of architecture is used, while the semantic segmentation branch is augmented with a Pyramid Pooling Module. Results for this method are submitted to the COCO and Mapillary Joint Recognition Challenge 2018. Our approach achieves a PQ score of 17.6 on the Mapillary Vistas validation set and 27.2 on the COCO test-dev set.
\end{abstract}

%%%%%%%%% BODY TEXT
\section{Introduction}
A key task in computer vision is image recognition, for which the ultimate goal is to recognize all elements in an image. At a high level these elements can be divided into two categories: \textit{things} and \textit{stuff} \cite{Forsyth1996}. \textit{Things} are usually countable objects, such as vehicles, persons and furniture items. On the other hand, \textit{stuff} is the set of remanining elements, usually not countable, such as sky, road and water. Within image recognition, many tasks have been introduced to identify these elements in images. \textit{Instance segmentation} and \textit{semantic segmentation} are two of such tasks that have become very prominent, with state-of-the-art methods \cite{He2017, Liu2018} and \cite{Chen2018a, Zhao2017}, respectively. The first task, instance segmentation, focuses on the detection and segmentation of \textit{things}. If an object is detected, a pixel mask is predicted for this object, and the output of such a method is a set of pixel masks. By design, this method does not account for all elements in an image, as it does not consider \textit{stuff} classes. The second task, semantic segmentation, does consider all elements, as the aim is to make a class prediction for each pixel in an image, for both \textit{things} and \textit{stuff} classes. However, the semantic segmentation output does not differentiate between different instances of \textit{things} classes. As a result, both methods lack the ability to fully describe the contents of an image. 

\begin{figure}[t]
\centering
\includegraphics[width=0.49\linewidth]{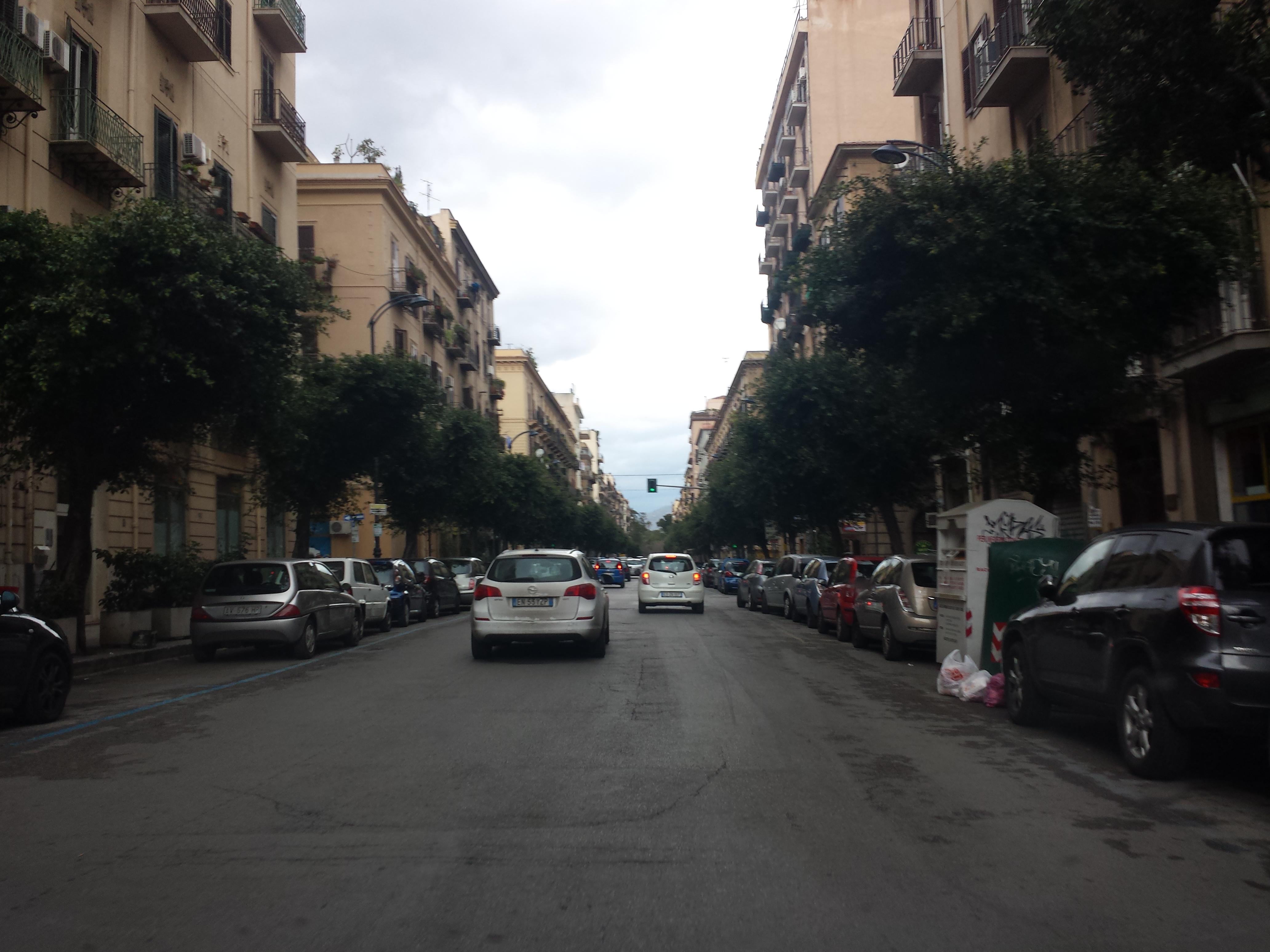}
\includegraphics[width=0.49\linewidth]{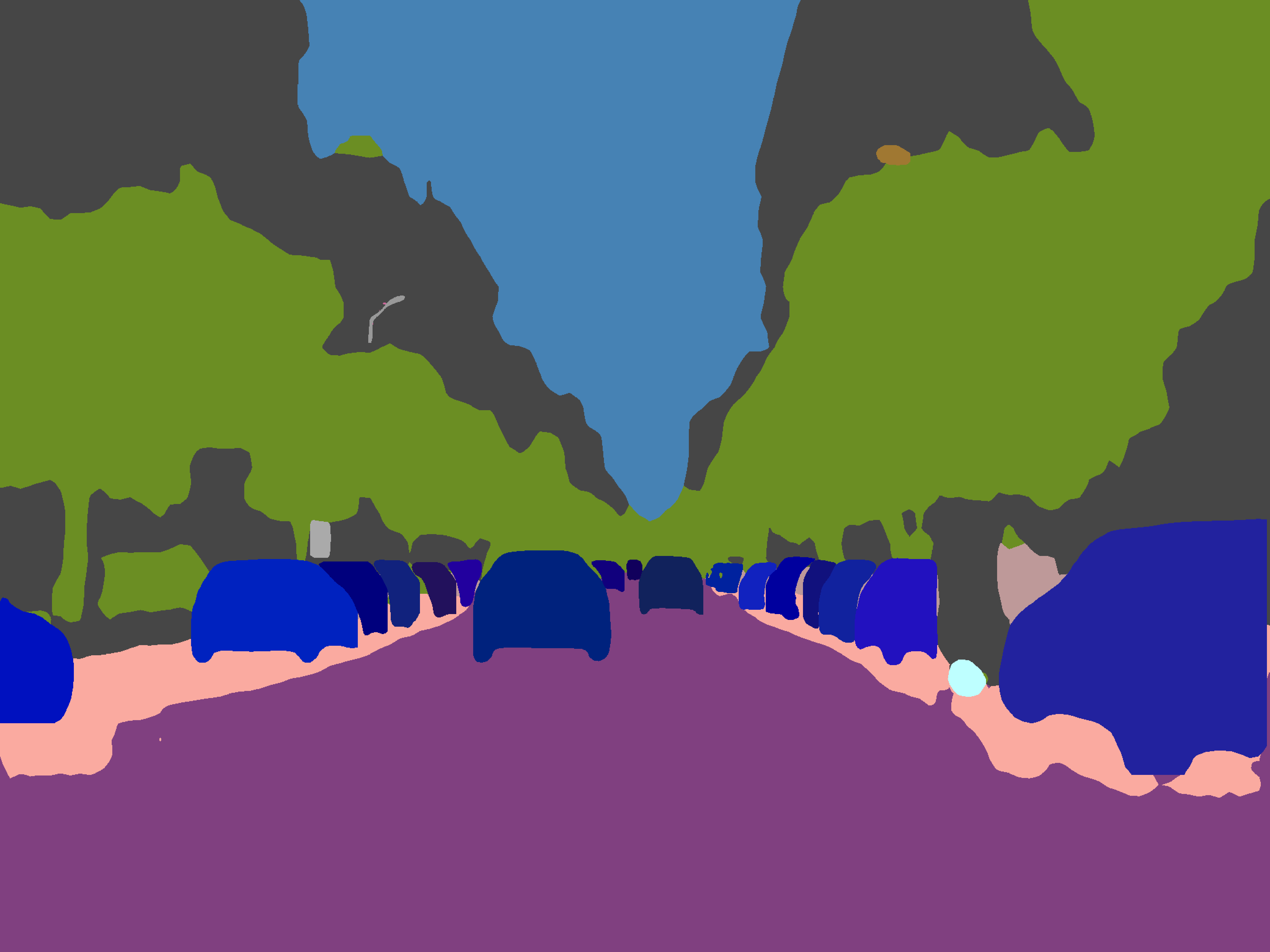}\\
\includegraphics[width=0.49\linewidth]{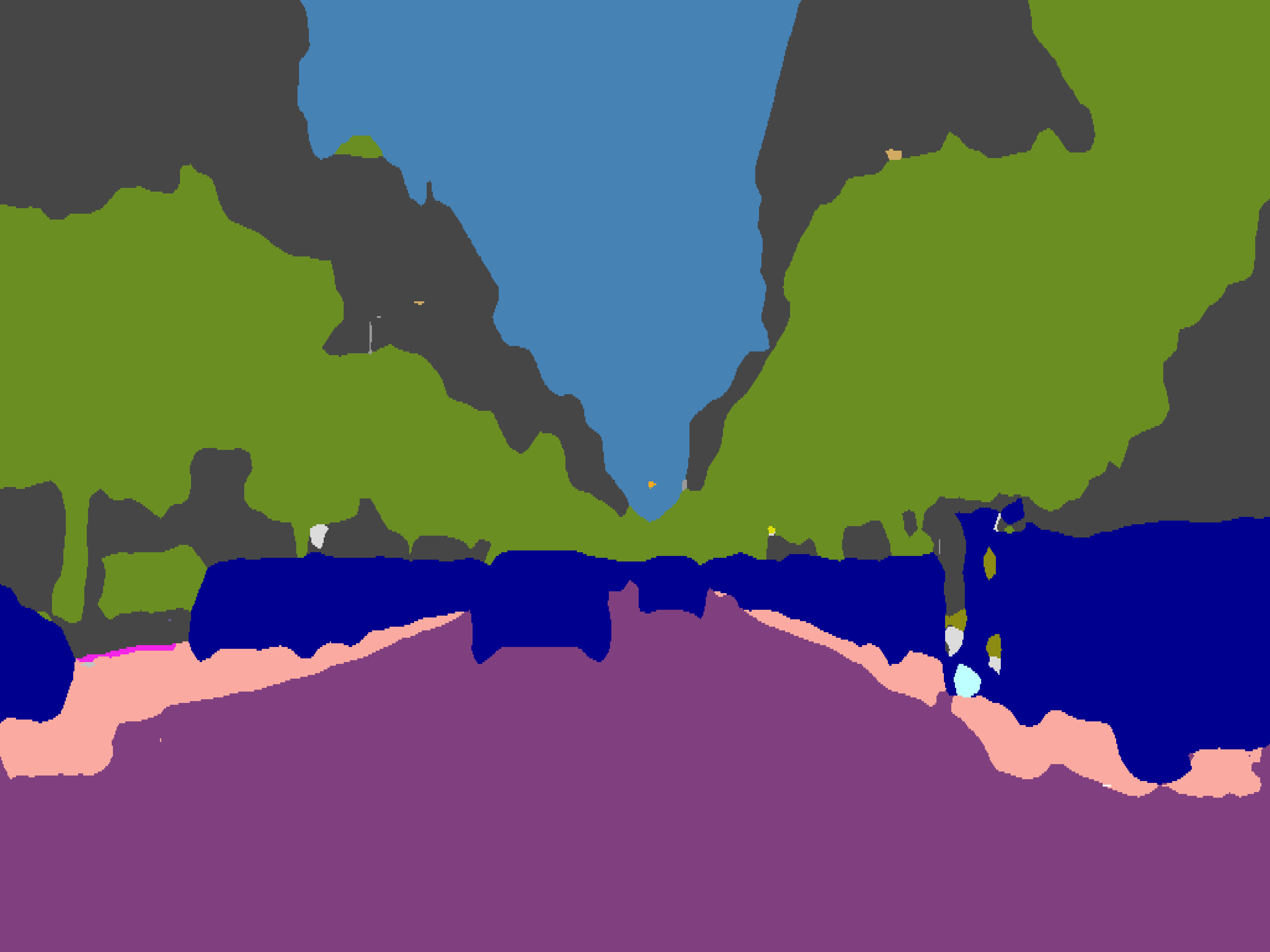}
\includegraphics[width=0.49\linewidth]{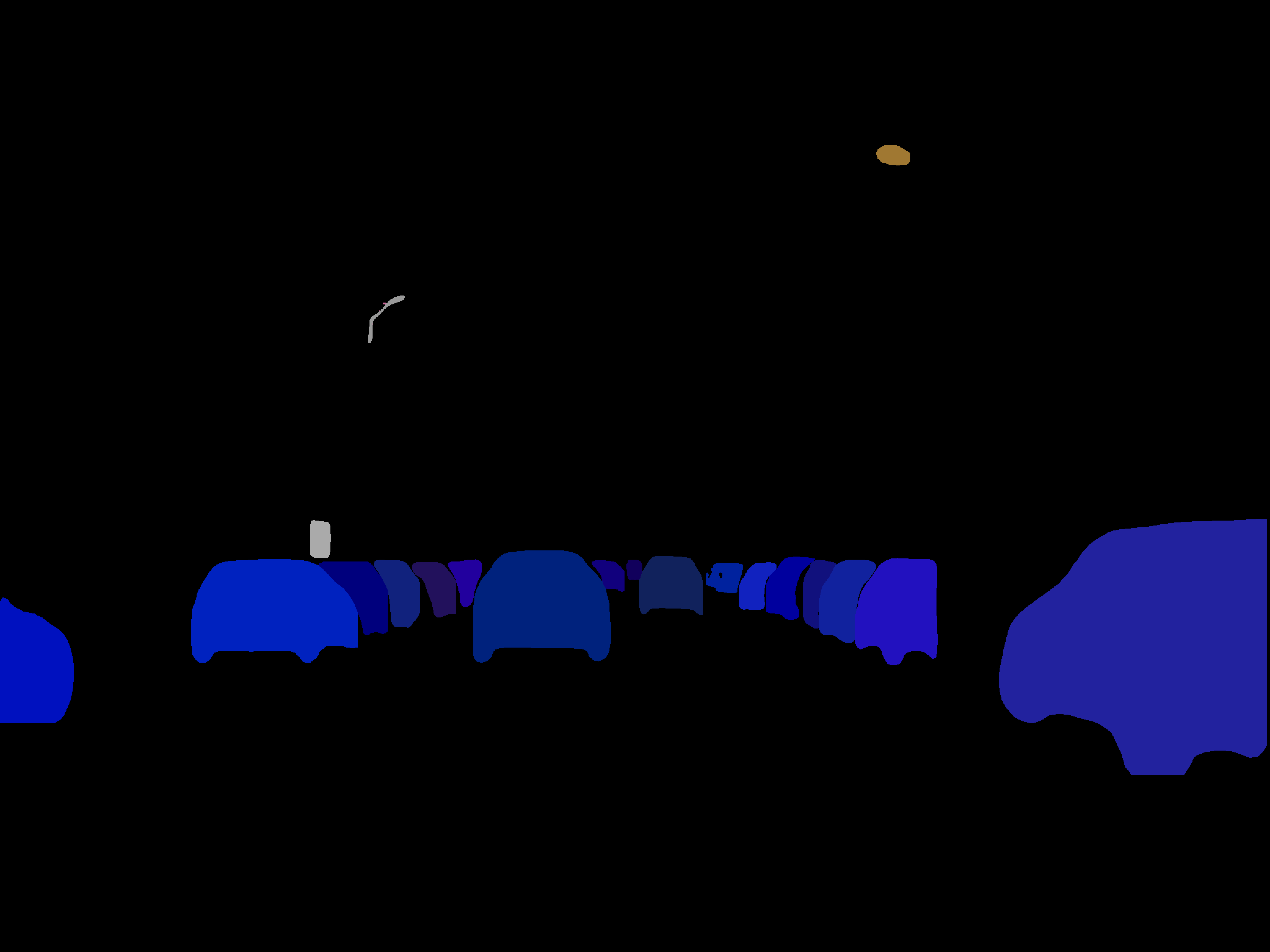}
\caption{Predictions by the network for an image from the Mapillary Vistas validation set. Top left: original image. Top right: panoptic segmentation. Bottom left: semantic segmentation. Bottom right: instance segmentation.}
\label{fig:intro_example_images}
\end{figure}

To fill this gap, the task of \textit{panoptic segmentation} is introduced in \cite{Kirillov2018}. For this task, each pixel of an image must be assigned with a class label and an instance \textit{id}. For \textit{things} classes, the instance id is used to distinguiscoh between different objects. On the other hand, for the \textit{stuff} classes, it is not necessary -- and sometimes not even possible -- to distinguish between different instances. Therefore, pixels in these classes always get the same instance id. In \cite{Kirillov2018}, a baseline method for this task is presented, for which they take the outputs of the best scoring instance segmentation and semantic segmentation networks, and combine them using basic heuristics to generate an output in the panoptic format.

Before the task of panoptic segmentation was formally introduced, there were already some publications that focused on exactly this task. In \cite{Uhrig2016}, depth layering and direction predictions are used to detect different instances of objects in a semantic segmentation map. In \cite{Arnab2017}, a Dynamically Instantiated Network is used to combine the outputs from an external object detector and an internal semantic segmentation network to form a panoptic-like output.

In this report, we present a single network that makes both instance segmentation and semantic segmentation predictions, using a shared feature extractor. These predictions are combined to form panoptic segmentation outputs using heuristics, augmenting those in \cite{Kirillov2018}. The main contribution of our approach is the fact that we apply a single network to make panoptic segmentation predictions.

%-------------------------------------------------------------------------
\section{Method}
We propose a Joint Semantic and Instance Segmentation Network (JSIS-Net) for panoptic segmentation. This method consists of two main sections: 1) a Convolutional Neural Network (CNN) that jointly predicts semantic segmentation and instance segmentation outputs (Section \ref{sec:network_architecture}) and 2) heuristics that are used to merge these outputs to generate panoptic segmentation predictions (Section \ref{sec:heuristics}).

\subsection{Network architecture}
\label{sec:network_architecture}
\begin{figure}[t]
\centering
\includegraphics[width=\linewidth]{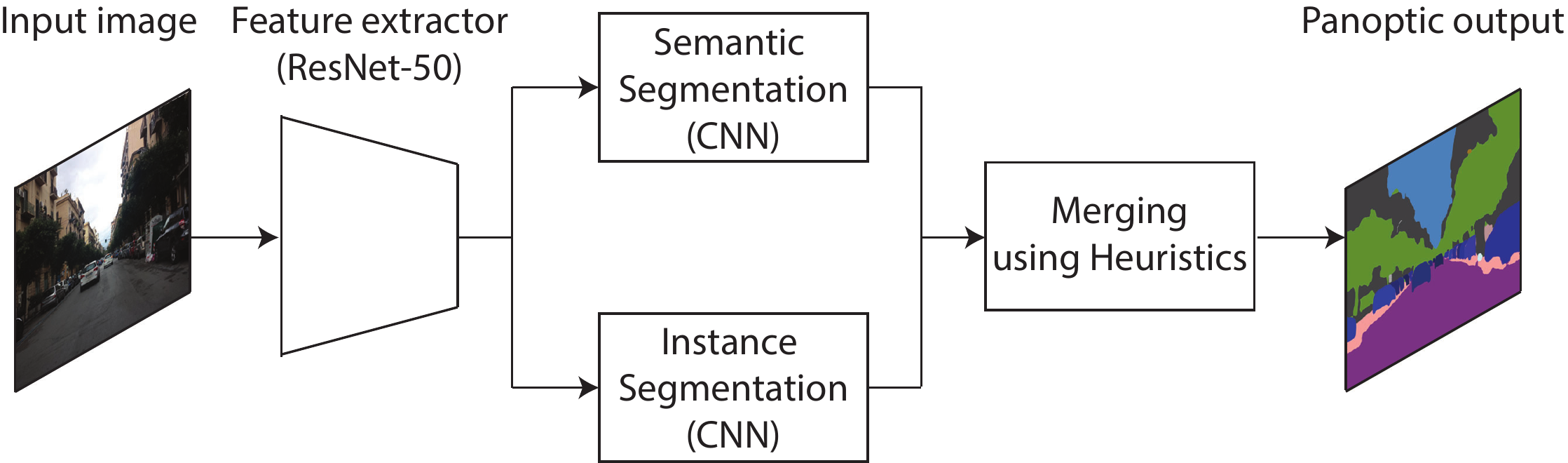}
\caption{The JSIS-Net architecture.}
\label{fig:architecture}
\end{figure}

We propose a CNN that jointly predicts semantic segmentation and instance segmentation outputs. The base of the network is a ResNet-50 feature extractor \cite{He2015}, which is shared by the semantic segmentation and instance segmentation branch. This is depicted in Figure \ref{fig:architecture}.

The semantic segmentation branch first applies a Pyramid Pooling Module to the generated feature map, as presented in \cite{Zhao2017}, and uses hybrid upsampling to reshape the predictions to the size of the input image \cite{Meletis2018}. This hybrid upsampling first applies a deconvolution operation and then bilinearly resizes the predictions to the dimensions of the input image. The output of this branch is a pixel map where each entry corresponds to the predicted class label for that pixel in the input image.

The instance segmentation branch is based on Mask R-CNN \cite{He2017}. First, a Region Proposal Network (RPN) is used to generate region proposals for potential objects in the image. The features corresponding to these proposals are then extracted from the feature map and subjected to the final layers of ResNet-50. Finally, these features are used to make three different parallel predictions: a classification score, bounding box coordinates, and an instance mask. After applying non-maximum suppression, the output of this branch is a set of pixel clusters with class labels predicted to correspond to the location of different objects in the image. With post-processing, these pixel clusters are transformed to form per-object normalized instance masks with the dimensions of the input image.

\subsubsection{Loss balancing for joint learning} 
To enable joint learning for this network, a single loss function is formed. This means that the various loss functions from the different network branches have to be combined and balanced. This total loss, $L_{tot}$ has the following form:

\begin{equation*}
\begin{aligned}
L_{tot} &= {\lambda_1}{L_{rpn,obj}} + {\lambda_2}{L_{rpn,reg}} \\ &+
{\lambda_3}{L_{det,cls}} + {\lambda_4}{L_{det,reg}} \\ &+
{\lambda_5}{L_{mask}} \\ &+ {\lambda_6}{L_{seg}} \\ &+  {\lambda_7}{R}.
\end{aligned}
\end{equation*}

Here, $L_{rpn,obj}$ is the softmax cross-entropy objectness loss function for the RPN, $L_{rpn,reg}$ is the smooth L1 regression loss function for the RPN \cite{Girshick2015}, $L_{det,cls}$ is the softmax cross-entropy classification loss function for object detection, $L_{det,reg}$ is the smooth L1 regression loss function for the object bounding boxes, $L_{mask}$ is the sigmoid cross-entropy loss on the instance masks, and $L_{seg}$ is the sparse softmax cross-entropy segmentation loss on the segmentation outputs. Finally, $R$ is the L2 regularization on the model parameters. The weights $\lambda_1...\lambda_n$ are the $n$ tuning parameters that are used to balance the losses.

\subsection{Merging heuristics}
\label{sec:heuristics}
Our network outputs semantic segmentation and instance segmentation predictions. These outputs cannot be directly combined into the panoptic segmentation output format. For panoptic segmentation, two values have to be predicted for each pixel: a class label and an instance id. There are essentially two conflicts that need to be solved before being able to generate this output: overlapping instance masks and conflicting predictions for \textit{things} classes.

\subsubsection{Overlapping instance masks}
Because the instance segmentation prediction is essentially based on an object detector and many overlapping region proposals, there is usually overlap between different predicted instance masks. This could be solved by applying non-maximum suppression to all overlapping instance masks, but this would remove many \textit{true} predictions. Instead, we chose to leverage the per-instance probability maps to resolve conflicting sections. In the case that two or more instance masks predict that a certain pixel belongs to \textit{their} object, the pixel is assigned to the instance mask with the highest probability at that specific pixel. These probabilities are predicted by the instance segmentation branch for each pixel in an instance mask. As a result of this heuristic, all output pixels are assigned to only one object.

\subsubsection{Conflicting predictions for \textit{things} classes}
Unlike the \textit{stuff} classes, which are only considered in the semantic segmentation branch, the \textit{things} classes are part of the prediction of both the semantic segmentation and the instance segmentation branch. As a result, there are inevitably \textit{things} prediction conflicts between the two outputs. Because the semantic segmentation output does not distinguish between different instances of objects, the two outputs cannot directly be compared. For this reason, we opt for the merging heuristics used in \cite{Kirillov2018}: we remove all \textit{things} classes from the semantic segmentation output, and replace them with the most probable \textit{stuff} class according to the semantic segmentation prediction. This leads to a segmentation map consisting of only \textit{stuff} class labels. Subsequently, the instance segmentation output is used to replace \textit{stuff} predictions at pixels where \textit{things} are predicted. Hence, the instance segmentation output is prioritized over the semantic segmentation output. As a second heuristic, any predicted \textit{stuff} class with a total pixel count below 4096 is removed from the output as well, and replaced by the next best scoring \textit{stuff} class above this threshold. This is done because it is very unlikely that a \textit{stuff} class consists of such a limited number of pixels, if it is present in an image.

After resolving these conflicts, all detected objects get a unique id, and the network has the desired output: a class label and an instance id for all pixels in the input image.

%------------------------------------------------------------------------

\section{Implementation}

\begin{table*}[ht]
\centering
\begin{tabular}{ l || c | c | c | c | c | c | c | c | c }
Dataset & PQ & SQ & RQ & PQ\textsuperscript{Th} & SQ\textsuperscript{Th} & RQ\textsuperscript{Th} & PQ\textsuperscript{St} & SQ\textsuperscript{St} & RQ\textsuperscript{St} \\
\hline
Mapillary Vistas \texttt{val} & 17.6 & 55.9 & 23.5 & 10.0 & 47.6 & 14.1 & 27.5 & 66.9 & 35.8 \\
COCO \texttt{test-dev} & 27.2 & 71.9 & 35.9 & 29.6 & 71.6 & 39.4 & 23.4 & 72.3 & 30.6 \\
COCO \texttt{val} & 26.9 & 72.4 & 35.7 & 29.3 & 72.1 & 39.2 & 23.3 & 73.0 & 30.4 \\
\end{tabular}
\caption{The best overall results on the COCO and Mapillary Vistas datasets. St = \textit{stuff} classes; Th = \textit{things} classes.}
\label{tab:overall_results}
\end{table*}

During training, the entire network is trained jointly, using a stochastic gradient descent optimizer with a momentum of 0.9. The initial learning rate is 0.075, and the learning rate is decreased twice with a factor of 2. The time steps at which this happens depend on the dataset that is trained on. The loss and regularization weights are provided in Table \ref{tab:loss_weights}. The network is initialized using weights pre-trained on the ImageNet dataset \cite{Deng2009}, and it is always trained on a single Titan Xp GPU. All presented results are from a single model.

\begin{table}[H]
\centering
\begin{tabular}{ c | c | c | c | c | c | c}
$\lambda_1$ & $\lambda_2$ & $\lambda_3$ & $\lambda_4$ & $\lambda_5$ & $\lambda_6$ & $\lambda_7$ \\ \hline
1.0 & 1.0 & 1.0 & 0.15 & 0.3 & 1.0 & 5.5e-5 \\
\end{tabular}
\caption{The loss and regularization weights.}
\label{tab:loss_weights}
\end{table}

The network is trained and applied to two different datasets: Mapillary Vistas \cite{Neuhold2017} and COCO \cite{Lin2014}. For both datasets we use slightly different hyperparameters.

\textbf{Mapillary Vistas.} For Mapillary Vistas, the feature extractor has input dimensions of 512 x 1024 pixels, and we use a batch size of 2. All input images are resized to these dimensions before being fed to the network. The network is trained for 19 epochs, and the learning rate is decreased after 7 and 13 epochs. This dataset consists of three splits: \textit{training}, \textit{validation} and \textit{testing}. The network is trained on the training set and the hyperparameters are tuned by evaluating on the validation set. In this report, the performance on the validation set is presented. The performance on the test set will be known when the results of the Mapillary Vistas Panoptic Segmentation Challenge are published.

\textbf{COCO.} Because the COCO images are much smaller than the Mapillary Vistas images, the input dimensions of the feature extractor are decreased to 400 x 400 pixels. Again, all input images are resized to these dimensions. In this case, a batch size of 5 is used. Because the amount of training images is much larger for COCO, the network is trained for 8 epochs, and the learning rate is decreased after 4 and 7 epochs. This dataset consists of four splits: \textit{training}, \textit{validation}, \textit{test-dev} and \textit{test-challenge}. The network is trained on the training set and the hyperparameters are tuned by evaluating on the validation set. In this report, the performance on both the validation and the test-dev set is presented. The performance on the test-challenge set will be known when the results of the COCO Panoptic Segmentation Challenge are published.

%------------------------------------------------------------------------
\section{Results}

The results on the aformentioned datasets have been submitted to the COCO and Mapillary Joint Recognition Challenge 2018. At the time of publication, the results on the challenge test sets for both datasets have not yet been announced.

\begin{figure}[t]
\centering
\includegraphics[width=0.49\linewidth]{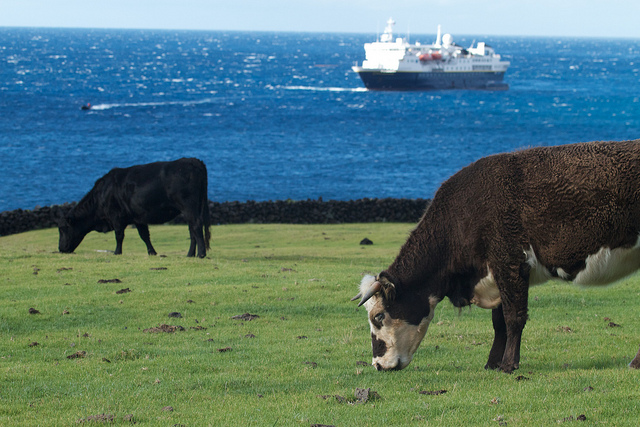}
\includegraphics[width=0.49\linewidth]{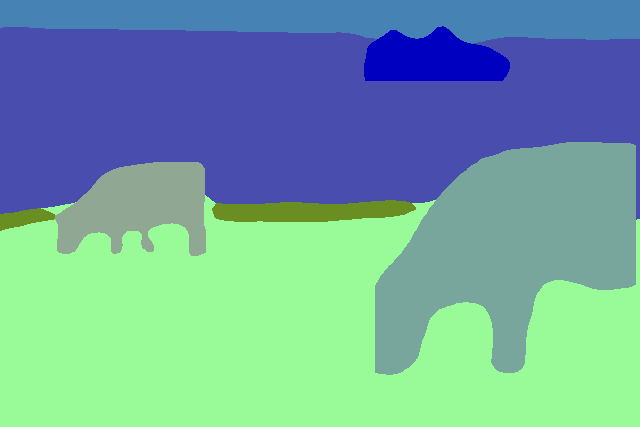}\\
\includegraphics[width=0.49\linewidth]{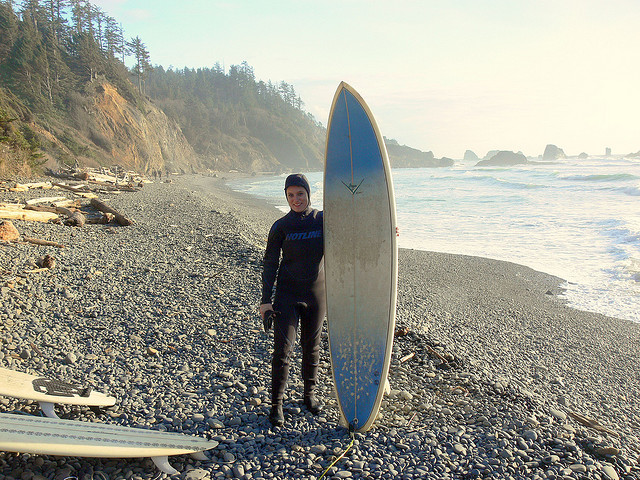}
\includegraphics[width=0.49\linewidth]{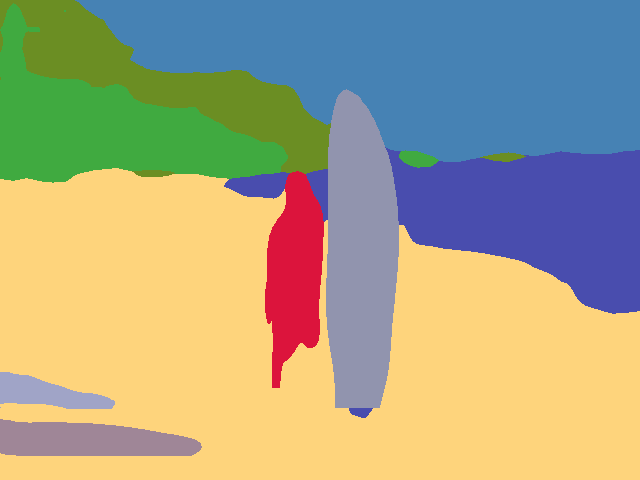}\\
\includegraphics[width=0.49\linewidth]{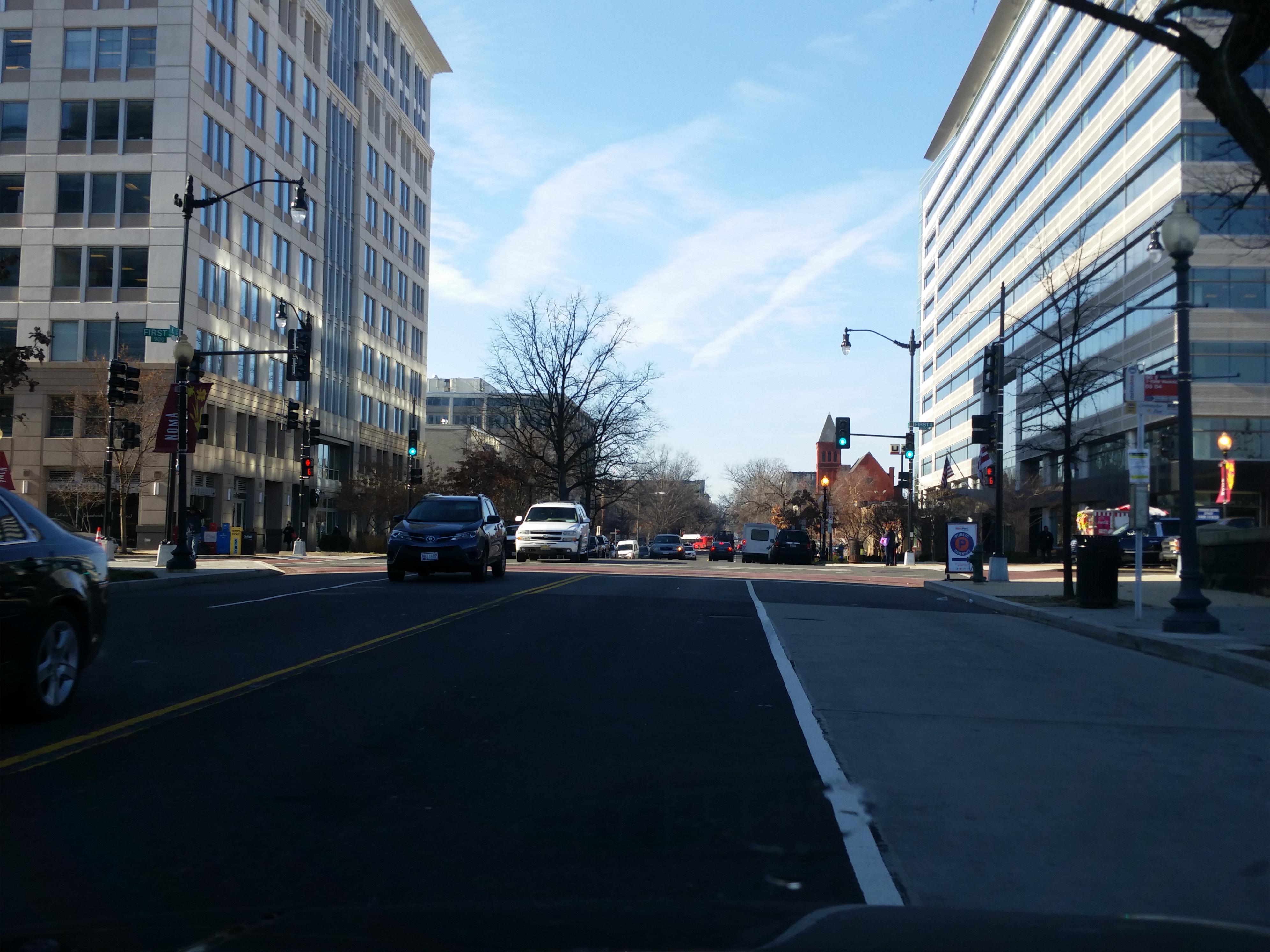}
\includegraphics[width=0.49\linewidth]{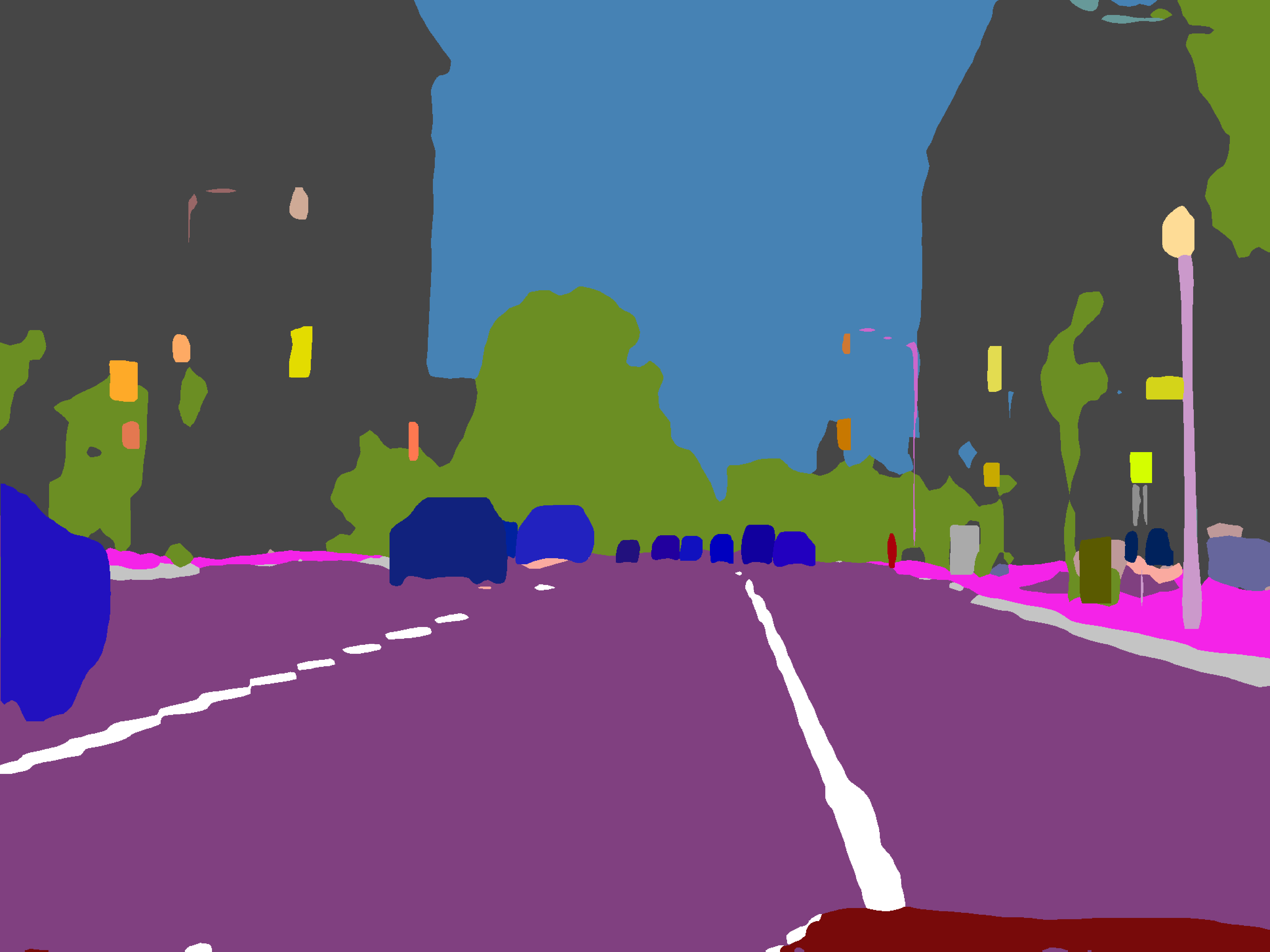}\\
\includegraphics[width=0.49\linewidth]{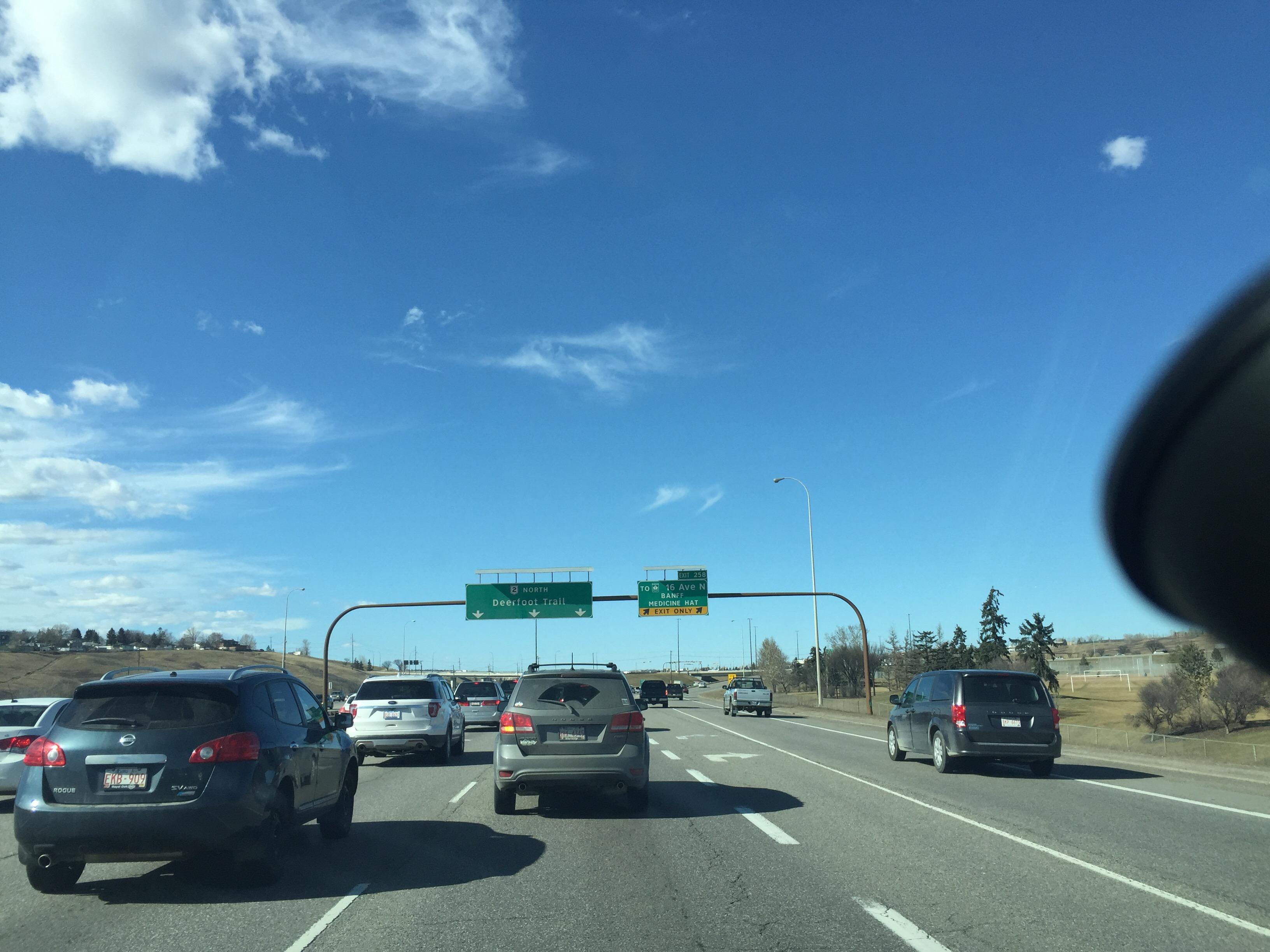}
\includegraphics[width=0.49\linewidth]{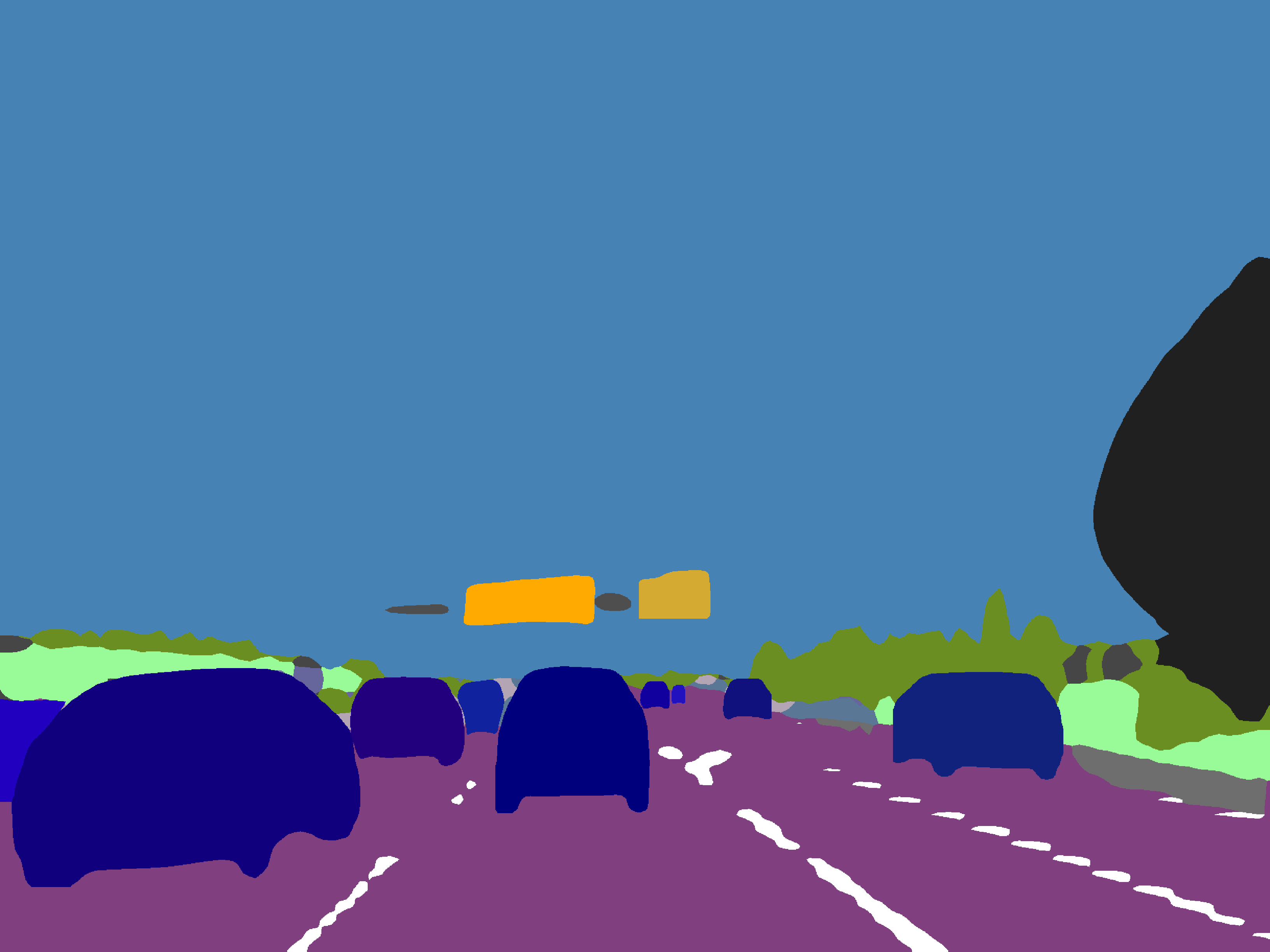}\\
\includegraphics[width=0.49\linewidth]{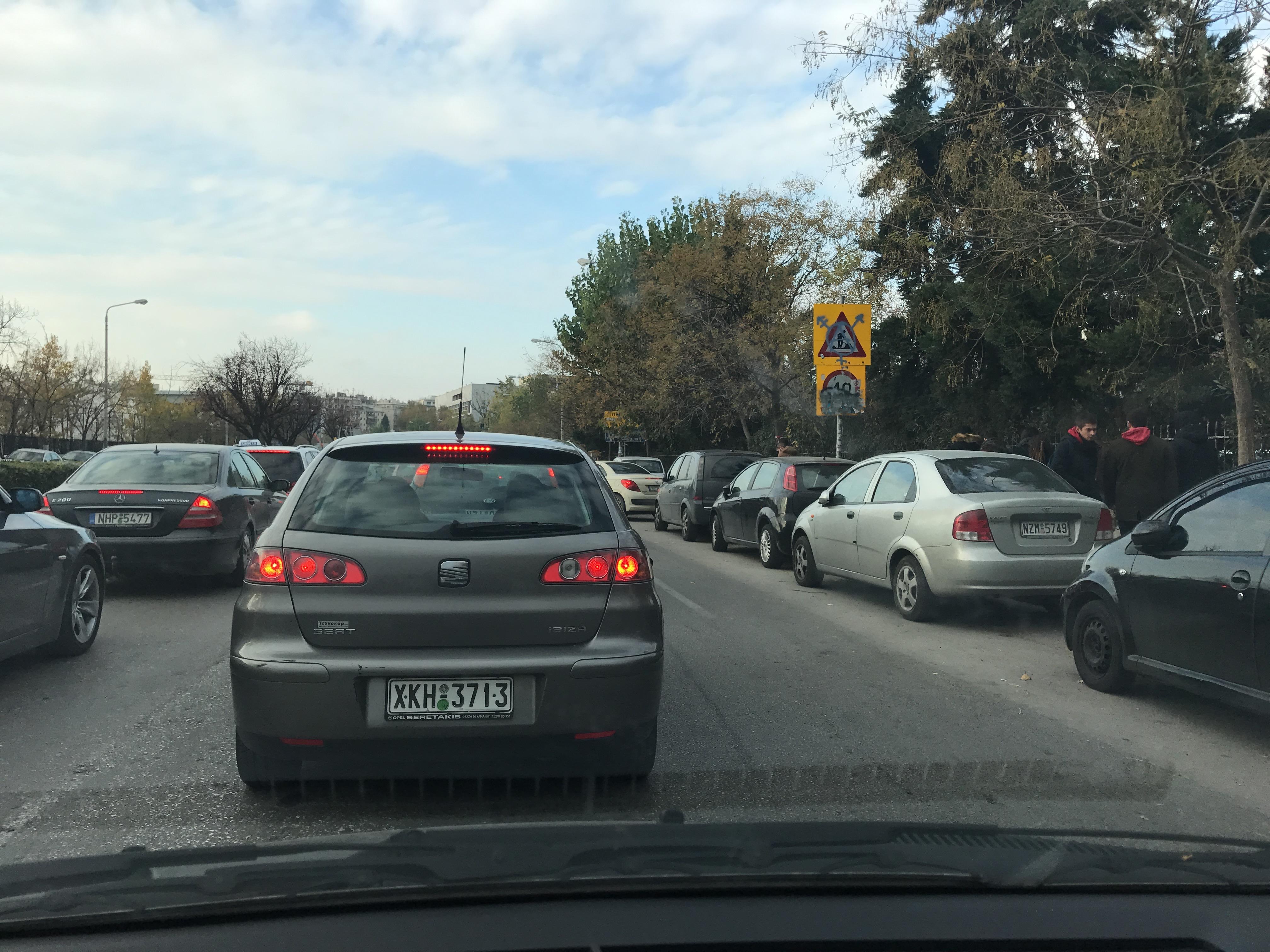}
\includegraphics[width=0.49\linewidth]{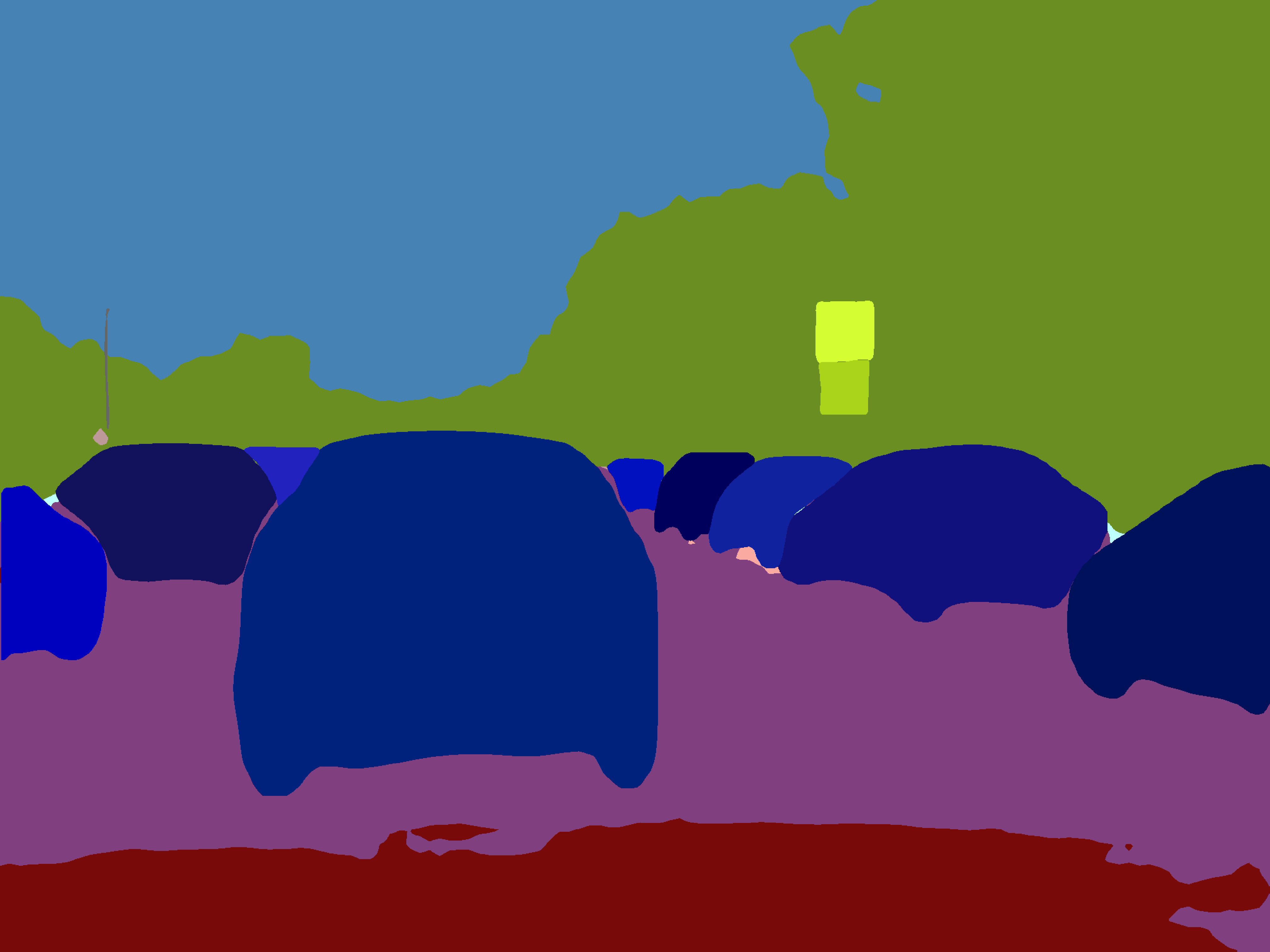}
\caption{Panoptic segmentation predictions by the network. Images are from the COCO test-dev set (first two rows) and the Mapillary Vistas test set (last three rows).} \label{fig:more_example_images}
\end{figure}

To enable proper evaluation of panoptic segmentation methods, a new metric is introduced as a main evaluation criterium for the challenge, called Panoptic Quality (PQ) \cite{Kirillov2018}. This metric is designed to assess both the segmentation and recognition performance of the different methods, and it can be divided into the Segmentation Quality (SQ) and the Recognition Quality (RQ). The best overall results on the two datasets are shown in Table \ref{tab:overall_results}. Examples of model outputs can be seen in Figures \ref{fig:intro_example_images} and \ref{fig:more_example_images}. Especially in Figure \ref{fig:intro_example_images}, it can clearly be seen that the predictions by both methods are combined to generate predictions for every pixel in the input image, while differentiating between different \textit{things}.

\subsection{Joint training vs. independent training}
\label{sec:joint_training}
The major difference between our method and the baseline method proposed in \cite{Kirillov2018} is the fact that we jointly learn the semantic segmentation and instance segmentation branch, instead of using two independently trained models. To evaluate the effectiveness of this joint approach, we compare joint training on our network with independent training of the instance segmentation and semantic segmentation branches of our network. We use the same hyperparameters for all training runs, we train on Mapillary Vistas for 17 epochs, and we evaluate on the Mapillary Vistas validation set. To generate a panoptic output, we merge the results from the independently trained networks using the same heuristics used for the joint approach.

As evaluation criteria, we use not only the PQ, but also metrics to assess the segmentation and instance segmentation performance. For semantic segmentation, we use the commonly used mean Intersection over Union (mIoU), and for instance segmentation we use the mean Average Precision at an IoU threshold of 0.5 (mAP\textsuperscript{0.5}). The results are provided in Table \ref{tab:joint_training}.

\begin{table}[ht]
\centering
\begin{tabular}{ l | c | c | c }
 & mIoU & mAP\textsuperscript{0.5} & PQ \\ \hline
Joint training & 34.7 & 8.4 & 17.4 \\
Semantic segmentation only & 33.6 & - & 15.0 \\
Instance segmentation only & - & 6.5 & 15.0 \\
\end{tabular}
\caption{The performance of the jointly trained model compared with independently trained models, evaluated on the Mapillary Vistas validation set.}
\label{tab:joint_training}
\end{table}

From the results it is clear that the jointly trained network outperforms the independently trained networks in all evaluated metrics. It should be noted that all experiments are conducted with the same learning rate and regularization weight. 

\subsection{Detecting small objects}
From the results in Table \ref{tab:overall_results}, it is clear that there is a large gap in PQ between \textit{things} and \textit{stuff} classes on the Mapillary Vistas dataset. When looking at the network output qualitatively, it appears that the instance segmentation branch of the network is not able to detect small and \textit{oddly-shaped} objects very well.

Since we suspect the cause for this problem to be a sub-optimal performing RPN, we evaluate the performance of the RPN on the two different datasets. We do this by assessing the mean recall, which is the mean of the recall values of all images in a given image set. Recall is defined as

\begin{equation*}
\begin{aligned}
recall = \frac{|TP|}{|TP|+|FN|},
\end{aligned}
\end{equation*}

where $|TP|$ is the number of true positives and $|FN|$ is the number of false negatives. We define a true positive as a ground-truth object bounding box that has an IoU $\geq$ 0.5 with a region proposal generated by the RPN. A false negative is a ground-truth object bounding box thas does not meet this requirement. Essentially, the recall is the fraction of object bounding boxes in the ground-truth that is covered by the region proposals generated by the RPN. 

\begin{table}[ht]
\centering
\begin{tabular}{ l | c }
& mean recall \\ \hline
COCO \texttt{val} & 0.827 \\
Mapillary Vistas \texttt{val} & 0.363 \\
\end{tabular}
\caption{The RPN mean recall of the model on COCO and Mapillary Vistas.}
\label{tab:recall}
\end{table}

The RPN mean recall for the two different datasets is given in Table \ref{tab:recall}. From this, it becomes clear that the mean recall for the Mapillary Vistas dataset is much lower than for COCO. On average, only 36.3\% of the objects is covered by the region proposals. As a result, it is nearly impossible for the remaining part of the instance segmentation branch to detect the majority of objects in an image. This indicates that the RPN is currently a bottleneck in our approach.

%-------------------------------------------------------------------------
\section{Conclusion}
We presented a method that is able to generate panoptic segmentation predictions by merging outputs from a Joint Semantic and Instance Segmentation Network. As is clear from Section \ref{sec:joint_training}, this joint approach has the potential to outperform independently trained networks.

Although this method works, the performance is worse than the baseline presented in \cite{Kirillov2018}. For this report, we have only tested the architecture with one feature extractor, instance segmentation method and semantic segmentation method. There is no reason, however, that this architecture would not work with other individual methods. Because of this, the performance of the method is greatly dependent on the performance and complexity of the individual sub-methods that are used, and the resources that are required to apply these methods. The hyperparameters that we used for training could also be sub-optimal. After submitting the results to the challenge, we were able to achieve better results for PQ, but the performance is still not as desired.

In future work, we want to further explore the potential benefits of joint learning for panoptic segmentation.

{\small
\bibliographystyle{ieee}
\bibliography{bibliography}

\begin{thebibliography}{10}\itemsep=-1pt

\bibitem{Arnab2017}
A.~Arnab and P.~H.~S. Torr.
\newblock {Pixelwise Instance Segmentation with a Dynamically Instantiated
  Network}.
\newblock In {\em 2017 IEEE Conference on Computer Vision and Pattern
  Recognition (CVPR)}, pages 879--888, July 2017.

\bibitem{Chen2018a}
L.~Chen, Y.~Zhu, G.~Papandreou, F.~Schroff, and H.~Adam.
\newblock {Encoder-Decoder with Atrous Separable Convolution for Semantic Image
  Segmentation}.
\newblock {\em arXiv preprint arXiv:1802.02611}, Feb. 2018.

\bibitem{Deng2009}
J.~Deng, W.~Dong, R.~Socher, L.~Li, K.~Li, and L.~Fei-Fei.
\newblock {ImageNet: A large-scale hierarchical image database}.
\newblock In {\em 2009 IEEE Conference on Computer Vision and Pattern
  Recognition}, pages 248--255, June 2009.

\bibitem{Forsyth1996}
D.~A. Forsyth, J.~Malik, M.~M. Fleck, H.~Greenspan, T.~Leung, S.~Belongie,
  C.~Carson, and C.~Bregler.
\newblock Finding pictures of objects in large collections of images.
\newblock In J.~Ponce, A.~Zisserman, and M.~Hebert, editors, {\em Object
  Representation in Computer Vision II}, pages 335--360, Berlin, Heidelberg,
  1996. Springer Berlin Heidelberg.

\bibitem{Girshick2015}
R.~Girshick.
\newblock {Fast R-CNN}.
\newblock In {\em 2015 IEEE International Conference on Computer Vision
  (ICCV)}, pages 1440--1448, Dec. 2015.

\bibitem{He2017}
K.~He, G.~Gkioxari, P.~Doll\'{a}r, and R.~Girshick.
\newblock {Mask R-CNN}.
\newblock In {\em 2017 IEEE International Conference on Computer Vision
  (ICCV)}, pages 2980--2988, Oct 2017.

\bibitem{He2015}
K.~He, X.~Zhang, S.~Ren, and J.~Sun.
\newblock {Deep Residual Learning for Image Recognition}.
\newblock In {\em 2016 IEEE Conference on Computer Vision and Pattern
  Recognition (CVPR)}, pages 770--778, June 2016.

\bibitem{Kirillov2018}
A.~Kirillov, K.~He, R.~Girshick, C.~Rother, and P.~Doll{\'a}r.
\newblock {Panoptic Segmentation}.
\newblock {\em arXiv preprint arXiv:1801.00868}, Jan. 2018.

\bibitem{Lin2014}
T.-Y. Lin, M.~Maire, S.~Belongie, J.~Hays, P.~Perona, D.~Ramanan,
  P.~Doll{\'a}r, and C.~L. Zitnick.
\newblock {Microsoft COCO: Common Objects in Context}.
\newblock In D.~Fleet, T.~Pajdla, B.~Schiele, and T.~Tuytelaars, editors, {\em
  Computer Vision -- ECCV 2014}, pages 740--755, Cham, 2014. Springer
  International Publishing.

\bibitem{Liu2018}
S.~{Liu}, L.~{Qi}, H.~{Qin}, J.~{Shi}, and J.~{Jia}.
\newblock {Path Aggregation Network for Instance Segmentation}.
\newblock {\em arXiv preprint arXiv:1803.01534}, Mar. 2018.

\bibitem{Meletis2018}
P.~Meletis and G.~Dubbelman.
\newblock {Training of Convolutional Networks on Multiple Heterogeneous
  Datasets for Street Scene Semantic Segmentation}.
\newblock {\em arXiv preprint arXiv:1803.05675}, Mar. 2018.

\bibitem{Neuhold2017}
G.~Neuhold, T.~Ollmann, S.~R. Bul{\`o}, and P.~Kontschieder.
\newblock {The Mapillary Vistas Dataset for Semantic Understanding of Street
  Scenes}.
\newblock In {\em 2017 IEEE International Conference on Computer Vision
  (ICCV)}, pages 5000--5009, Oct. 2017.

\bibitem{Uhrig2016}
J.~Uhrig, M.~Cordts, U.~Franke, and T.~Brox.
\newblock {Pixel-Level Encoding and Depth Layering for Instance-Level Semantic
  Labeling}.
\newblock {\em arXiv preprint arXiv:1604.05096}, Apr. 2016.

\bibitem{Zhao2017}
H.~Zhao, J.~Shi, X.~Qi, X.~Wang, and J.~Jia.
\newblock {Pyramid Scene Parsing Network}.
\newblock In {\em 2017 IEEE Conference on Computer Vision and Pattern
  Recognition (CVPR)}, pages 6230--6239, July 2017.

\end{thebibliography}
}

\end{document}